\documentclass[runningheads]{llncs}

\usepackage[T1]{fontenc}
\usepackage{graphicx}
\usepackage{xurl}
\usepackage{hyperref}
\usepackage{marvosym}
\usepackage{enumitem}
\usepackage{orcidlink}

\newcommand{\corrauth}{\textsuperscript{(\Letter)}}
\renewcommand{\orcidID}[1]{\orcidlink{#1}}
\begin{document}
\title{AniPrO: Interpretable Anime Image Provenance Detection via Multi-Dimensional Semantic Reasoning}
\titlerunning{AniPrO: Interpretable Anime Image Provenance Detection}
%
\author{Yan Liu\inst{1}\corrauth\orcidID{0009-0009-6813-7840} \and
Baoxiang Huang\inst{1}\orcidID{0009-0005-8956-7533} \and
Zi'an Wang\inst{2} \and
Wenbo Xie\inst{1}}
\authorrunning{Y. Liu et al.}
%
\institute{
School of Computer Science and Technology, Tongji University, Shanghai, China\\
\email{\{y\_an,2351753\}@tongji.edu.cn}
\and
College of Electronic and Information Engineering, Tongji University, Shanghai, China
}
\maketitle              
\begingroup
\renewcommand{\thefootnote}{}
\footnotetext{Accepted at the Computer Graphics International (CGI) 2026.}
\endgroup
\begin{abstract}
As generative AI becomes increasingly used in anime-style image creation, distinguishing human-drawn, AI-inpainted, and text-to-image images is important for copyright attribution, visual provenance, and content governance. Existing AI-generated image detectors mainly target real-world photographs and often overlook anime-specific cues such as flat coloring, exaggerated structures, and artistic line control. To address this gap, we propose AniPrO, a multi-dimensional description-enhanced framework for interpretable anime image provenance. Built upon AnimeDL-\(2\mathrm{M}\), AniPrO contains \(15{,}000\) balanced samples from a \(35{,}000\)-image candidate pool, covering Real, Inpainting, and Text2Image categories with structured five-dimensional descriptions. We further introduce AniPrO-SFD-Bench and AniPrO-MFR-Bench to evaluate provenance detection from statistical feature discrimination and multimodal fusion reasoning perspectives. Experiments show that structured semantic guidance reveals systematic AI-generation biases, such as the gap between global visual plausibility and local detail coherence, and improves the detection of challenging inpainting samples. The dataset and code will be released at: \url{https://github.com/YAN-LIU05/AniPrO}.

\end{abstract}
\section{Introduction}
\begin{figure}[t]
    \centering
    \includegraphics[width=\linewidth]{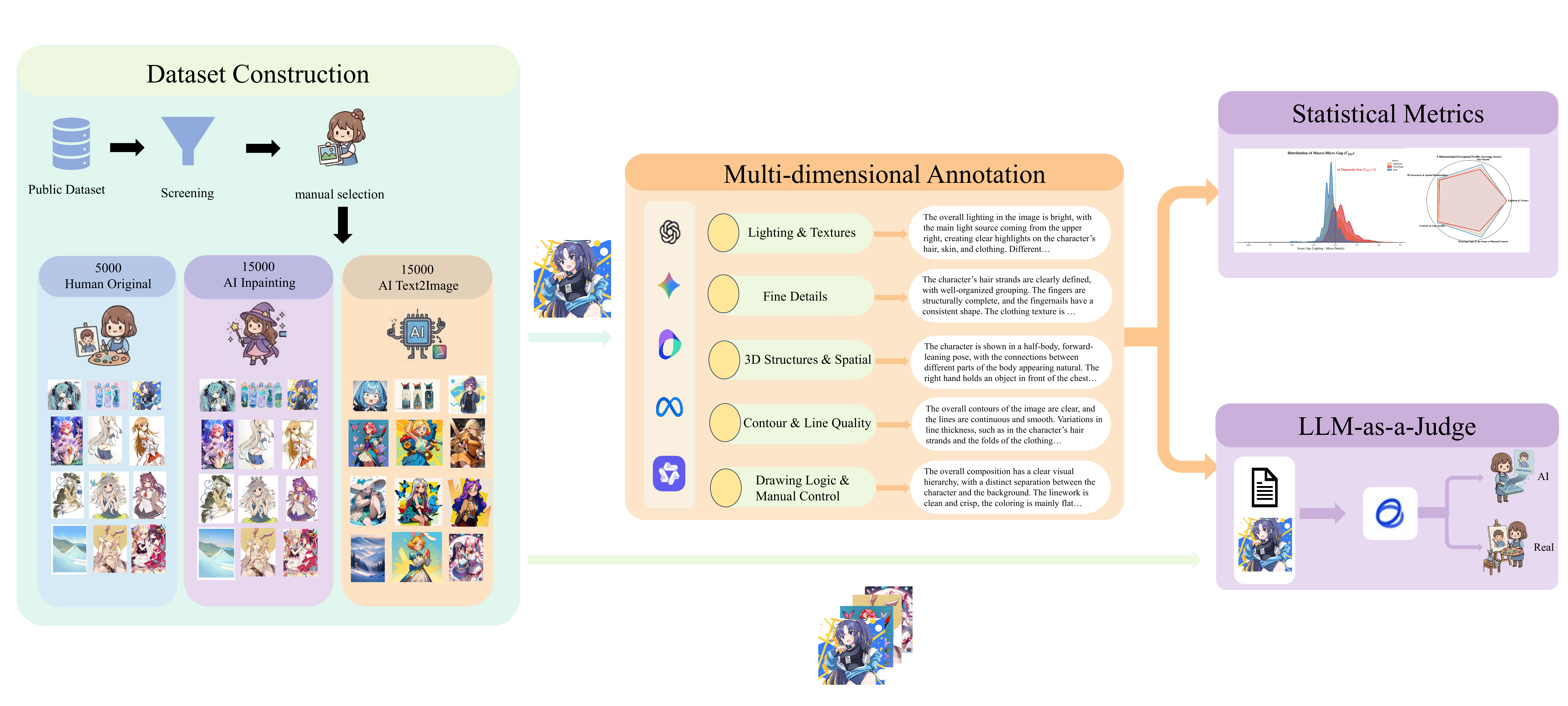}
    \caption{Overview of the proposed multi-dimensional description-enhanced framework. The pipeline consists of three main stages: 
    (1) Dataset Construction of the AniPrO dataset; 
    (2) Multi-dimensional Annotation, which extracts five-dimensional structured scores and texts using multimodal LLMs; and 
    (3) A dual-track evaluation system utilizing Statistical Metrics and LLM-as-a-Judge for final AI vs. Real image discrimination.}
    \label{fig:framework}
\end{figure}

Recent text-to-image and image-editing models, such as Stable Diffusion~\cite{stable_diffusion} and Midjourney~\cite{midjourney}, have made anime-style image generation increasingly accessible. These models can synthesize complete artworks from prompts and perform local inpainting, style transfer, and fine-grained editing, raising new challenges for copyright attribution, visual provenance, and AIGC governance. In this paper, we study anime image provenance detection, aiming to distinguish human-drawn artworks from AI-inpainted and text-to-image generated samples.

Existing AI-generated image detectors are mainly designed for real-world photographs and often rely on cues such as GAN artifacts, frequency anomalies, and facial distortions~\cite{ref3,ref4}. However, these cues are less reliable for anime images, where simplified line art, flat colors, non-photorealistic shading, exaggerated anatomy, and intentional perspective distortion are common artistic conventions. As a result, visual irregularities may indicate either AI generation or deliberate human stylization. This makes anime provenance detection not only a classification problem, but also an interpretability problem: a detector should expose which visual evidence supports its decision.

To address this issue, we construct AniPrO based on AnimeDL-\(2\mathrm{M}\)~\cite{ref5}, a balanced benchmark covering human-drawn, AI-inpainted, and text-to-image anime samples. Each image is annotated with structured descriptions along five anime-oriented semantic dimensions: lighting and textures, fine details, \(3\mathrm{D}\) structures and spatial relationships, contour and line quality, and drawing logic and manual control. Based on these annotations, we propose a multi-dimensional description-enhanced framework for interpretable anime provenance reasoning.

We further design two complementary evaluation tracks. \textbf{AniPrO-SFD-Bench} evaluates whether statistical patterns in multidimensional scores can discriminate \texttt{Real} and \texttt{AI} samples, while \textbf{AniPrO-MFR-Bench} evaluates whether structured descriptions improve multimodal fusion reasoning. Together, these two tracks assess the proposed framework from both statistical and semantic perspectives.

The main contributions are as follows:
\begin{enumerate}[label=\arabic*), leftmargin=*, itemsep=0pt, topsep=4pt]
    \item We construct AniPrO, a balanced \(15{,}000\)-image benchmark for anime image provenance with structured five-dimensional descriptions.
    \item We propose a five-dimensional interpretable reasoning framework tailored to anime-style visual provenance.
    \item We design a dual-track evaluation scheme combining statistical feature discrimination and multimodal semantic reasoning.
\end{enumerate}

\section{Related Work}

\subsection{AI-Generated Image Detection}

AI-generated image detection has been widely studied in digital image forensics, especially with the rise of diffusion models. Existing methods detect synthetic images using explicit artifacts such as abnormal textures and frequency inconsistencies~\cite{ref4}, or implicit cues such as reconstruction consistency, physical constraints, and cross-model generalization~\cite{ref6}. Recent vision-language and multimodal large language models further reformulate detection as visual question answering, description generation, or language-assisted classification~\cite{ref7}. However, most existing methods still emphasize authenticity prediction, while fine-grained provenance cues and interpretable reasoning remain insufficiently explored.

\subsection{Anime Image Analysis and Provenance}

Anime images differ from natural photographs in line art, flat coloring, shadow organization, and stylized deformation. Such visual irregularities may reflect either AI artifacts or deliberate artistic choices, making natural-image detectors difficult to transfer to anime-style images~\cite{ref5}. Existing anime datasets such as Manga109 and Danbooru2024~\cite{ref8} support retrieval, character recognition, pose estimation, and style analysis, but fine-grained anime provenance among human-drawn, AI-generated, and AI-inpainted images remains underexplored.

\subsection{AI-Generated Image Detection Benchmarks}

Existing AI-generated image detection benchmarks mainly focus on real-vs-fake classification and generative attribution~\cite{ref9,ref10}. Although some provide annotations such as class labels, editing types, degradation conditions, or manipulation masks, their evaluation is still largely accuracy-oriented. This limitation is more evident in anime images, where stylized visual cues often require interpretation beyond low-level artifacts. Therefore, an anime-specific benchmark with interpretable multidimensional descriptions is needed to evaluate both detection performance and reasoning quality.

\section{Method}

\subsection{Overview of the Framework}

The framework converts anime provenance detection from direct \texttt{Real} or \texttt{AI} classification into structured multidimensional reasoning. Given an input image, a VLM first generates scores and textual descriptions along five semantic dimensions. These outputs are then used in two evaluation tracks: AniPrO-SFD-Bench, which analyzes statistical patterns in multidimensional scores, and AniPrO-MFR-Bench, which evaluates whether structured descriptions improve multimodal provenance reasoning. The two tracks jointly assess score-level regularities and semantic reasoning ability.

\begin{figure}[t]
    \centering
    \includegraphics[width=0.75\linewidth]{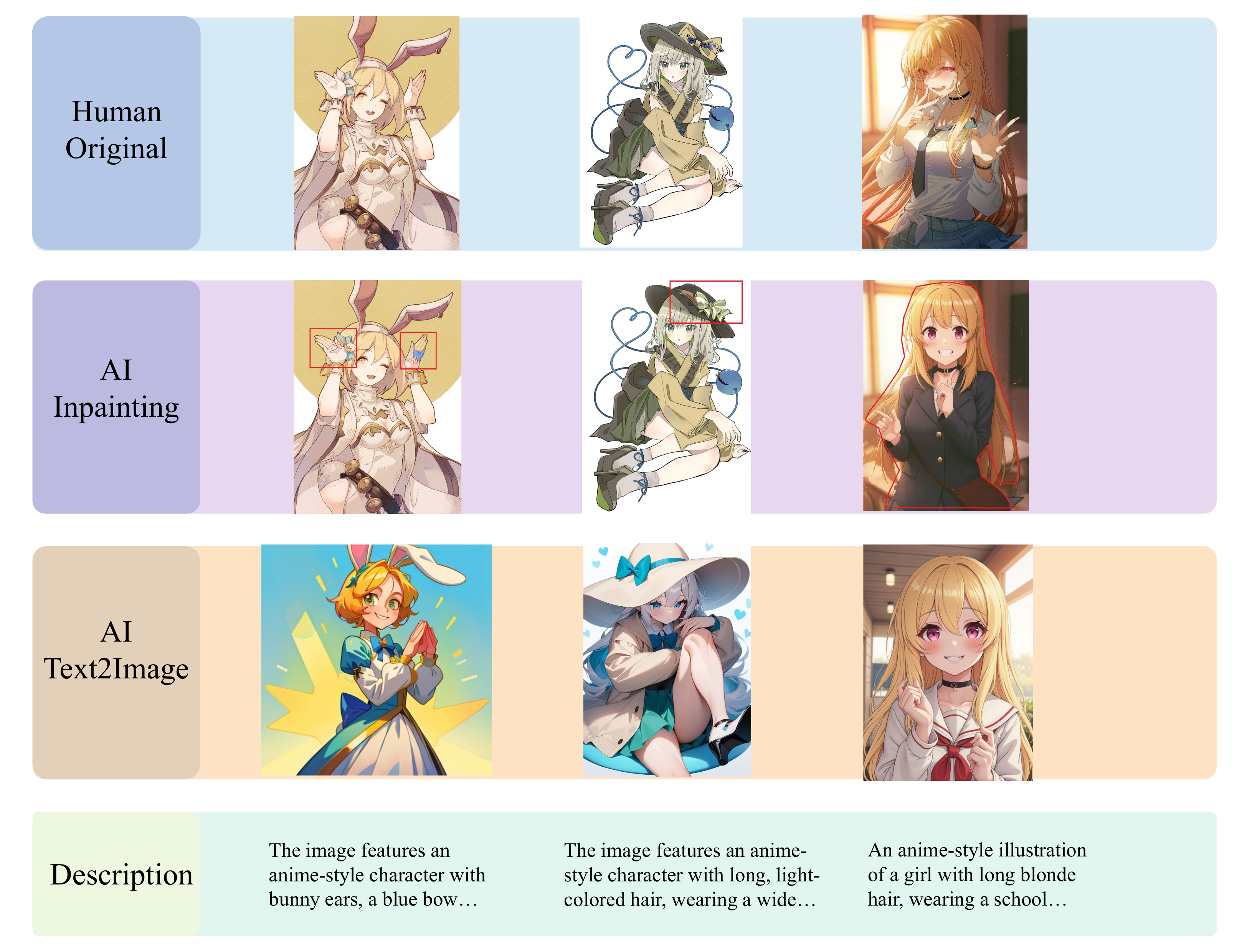}
    \caption{Representative Real, Inpainting, and Text2Image samples from AniPrO with structured descriptions.}
    \label{fig:samples}
\end{figure}

\subsection{Dataset Construction}

We construct AniPrO based on the publicly available AnimeDL-\(2\mathrm{M}\) dataset~\cite{ref5}, which provides anime images with provenance labels, including hand-drawn images (Real), AI-inpainting images (Inpainting), and text-to-image generated images (Text2Image). The AI-generated samples are produced by three mainstream diffusion models, including FLUX.1~\cite{ref11}, Stable Diffusion~\cite{stable_diffusion}, and Stable Diffusion XL~\cite{ref12}.

We manually select \(5{,}000\) high-quality Real images with complete compositions, clear subjects, and representative anime-style characteristics. For each selected Real image, we retain its corresponding Inpainting and Text2Image results, forming a \(35{,}000\)-image candidate pool. To build a balanced benchmark, we keep all Real images and randomly sample \(5{,}000\) images from each AI category, resulting in \(15{,}000\) images with \(5{,}000\) samples per class. Representative samples are shown in Fig.~\ref{fig:samples}. To reduce potential size-related bias, Real images are resized by setting the longer side to the median length while preserving the original aspect ratio.

AniPrO is designed as a balanced diagnostic benchmark rather than a large-scale training corpus. Its balanced class distribution supports controlled comparison among Real, Inpainting, and Text2Image samples, while cross-generator and cross-style generalization beyond FLUX.1, Stable Diffusion, and Stable Diffusion XL remains a direction for future expansion.

\subsection{Multi-dimensional Description and Prompt Design}

Instead of directly asking large language models to judge image authenticity, we prompt them to produce fine-grained descriptions and scores along five interpretable dimensions that reflect key visual cues of anime-style image naturalness.

\noindent\textbf{Lighting and Textures.}
This dimension evaluates tonal range, lighting consistency, shadows, highlights, and material rendering. AI-generated images may exhibit abnormal highlights, inconsistent shadows, or unnatural texture~\cite{ref13}.

\noindent\textbf{Fine Details.}
This dimension examines local shape coherence, edges, and detail organization. AI-generated images often contain excessive or chaotic details~\cite{ref13}, while human-drawn works show more deliberate and smoother arrangements~\cite{ref14}.

\noindent\textbf{Three-dimensional Structures and Spatial Relationships.}
This dimension evaluates spatial hierarchy, occlusion, perspective, object relationships, and body structure. AI-generated graphics may show spatial irregularities or unclear relationships~\cite{ref15}.

\noindent\textbf{Contour and Line Quality.}
This dimension focuses on line continuity, stability, and thickness variation. Human-drawn lines are generally purposeful and controlled~\cite{ref14}, whereas AI-generated lines may show breaks, drifting edges, clumping, or abrupt thickness changes~\cite{ref16}.

\noindent\textbf{Drawing Logic and Manual Control.}
This dimension evaluates layering, visual hierarchy, and intentional artistic control. Human-drawn images often show purposeful simplification and emphasis, while AI-generated images may show uneven detail distribution or inconsistent rendering logic.

\subsection{AniPrO-SFD-Bench}

AniPrO-SFD-Bench evaluates whether AI traces can be captured by statistical patterns in five-dimensional scores. We formulate provenance discrimination as binary classification, where Real images are labeled as \texttt{Real}, and both Inpainting and Text2Image samples are labeled as \texttt{AI}.

\subsubsection{Feature Construction.}
For each image \(i\), the five-dimensional scores corresponding to lighting, details, structure, lines, and manual control are denoted as:
\begin{equation}
    \mathbf{s}_i = \left[s_{i,1}, s_{i,2}, s_{i,3}, s_{i,4}, s_{i,5}\right].
    \label{eq:score_vector}
\end{equation}

Based on this multidimensional scoring representation, to characterize the systematic bias in the score distribution of AI images from different perspectives, four types of derived statistical features are further constructed for subsequent discriminant analysis.

The bucket effect captures the weakest dimension of an image:
\begin{equation}
    F_{\min} = \min\left(\mathbf{s}_i\right).
    \label{eq:f_min}
\end{equation}

The variance feature measures the degree of imbalance among the five dimensions, where \(\bar{s}_i\) is the mean score:
\begin{equation}
    F_{\mathrm{var}} = \frac{1}{4}\sum_{j=1}^{5}\left(s_{i,j}-\bar{s}_i\right)^2.
    \label{eq:f_var}
\end{equation}

The skewness feature describes the asymmetry of the score distribution, especially when only a few dimensions receive abnormally low scores:
\begin{equation}
    F_{\mathrm{skew}} =
    \frac{
    \frac{1}{5}\sum_{j=1}^{5}\left(s_{i,j}-\bar{s}_i\right)^3
    }{
    \left(
    \frac{1}{5}\sum_{j=1}^{5}\left(s_{i,j}-\bar{s}_i\right)^2
    \right)^{3/2}
    }.
    \label{eq:f_skew}
\end{equation}

The macro--micro gap measures the discrepancy between macro-level lighting consistency \(\left(s_{i,1}\right)\) and micro-level detail coherence \(\left(s_{i,2}\right)\), capturing the mismatch between overall visual plausibility and fine-grained local reliability:
\begin{equation}
    F_{\mathrm{gap}} = s_{i,1} - s_{i,2}.
    \label{eq:f_gap}
\end{equation}

\subsubsection{Single-feature Threshold Classification.}
To ensure balanced discrimination between real and AI-generated images, we adopt a unified criterion for threshold selection based on Youden's \(J\) index~\cite{ref17}:
\begin{equation}
    J = \mathrm{Recall}_{\mathrm{Real}} + \mathrm{Recall}_{\mathrm{AI}} - 1.
    \label{eq:youden_j}
\end{equation}

This objective explicitly balances the recognition performance of both classes and reduces the impact of data imbalance. For each feature, the decision direction is defined according to its statistical characteristics, enabling consistent and interpretable threshold-based classification.

\subsection{AniPrO-MFR-Bench}

AniPrO-MFR-Bench, namely the Multimodal Fusion Reasoning Benchmark, evaluates the comprehensive discrimination capabilities of multimodal large language models after integrating visual information with structured textual descriptions. Unlike the statistics-driven approach of AniPrO-SFD-Bench, this scheme adopts the ``LLM-as-a-Judge'' paradigm, in which the large language model directly acts as an expert in image source authentication, combining visual perception with semantic reasoning to make judgments.

In terms of task formulation, we also define source discrimination as a binary classification task between human-drawn images labeled as \texttt{Real} and AI-generated images labeled as \texttt{AI}, including Inpainting and Text2Image samples.

The two benchmarks validate the effectiveness of the proposed framework at both the statistical regularity and semantic reasoning levels. They complement each other and together constitute the comprehensive evaluation system of this paper.

\section{Experiment}
To validate the proposed multi-dimensional description-enhanced framework, we evaluate AniPrO from four aspects: statistical discrimination based on five-dimensional scores, multimodal provenance reasoning with structured descriptions, ablations of input modalities and semantic dimensions, and comparisons with supervised visual classifiers and human rater baselines.

\subsection{Experimental Setup}

All experiments are conducted on AniPrO under the binary setting in Sec.~3.4, where Real images are treated as Real and Text2Image/Inpainting samples as AI. We report Real Retention (Real Ret.), Text2Image Interception Rate (T2I Int.), Inpainting Interception Rate (Inpainting Int.), and Overall Accuracy (Overall Acc.). Real Ret. measures correctly retained human-drawn images, while the two interception rates measure correctly detected AI samples in each subtype.

For the feature extraction and semantic classification stages, we incorporate current mainstream vision-language models (VLMs) and large language models (LLMs), including Gemini-\(2.5\)-Pro~\cite{ref18}, GPT-\(5.2\)~\cite{ref19}, Doubao-\(1.5\)-vision-pro~\cite{ref20}, and GLM-\(4.5\)V~\cite{ref21}, as well as open-source models such as Qwen3-VL (235B/30B)~\cite{ref22} and Llama-\(4\) (Maverick/Scout)~\cite{ref23}, to verify the framework's generalizability across different base models.

During evaluation, less than 5\% of the images were excluded from the final quantitative analysis because they triggered API providers' safety filters, leading to refusals or empty responses.

\subsection{Quantitative Discrimination Experiment Based on Statistical Features}

This section examines whether AI-generated and inpainting images exhibit statistically detectable anomalies based solely on five-dimensional score distributions, without relying on complex semantic reasoning.

\subsubsection{Single-feature Threshold Discriminant Analysis.}
To validate the effectiveness of the constructed statistical features in the source discrimination task, we first evaluate the discriminative ability of each derived statistical feature on AniPrO-SFD-Bench. Using scores extracted by Gemini-\(2.5\)-Pro as an example, the univariate threshold results are shown in Table~\ref{tab:single_feature}.

\begin{table}
    \centering
    \caption{Univariate discriminatory performance based on derived statistical features using scores extracted from Gemini-\(2.5\)-Pro as an example.}
    \label{tab:single_feature}
    \begin{tabular}{lcccc}
        \hline
        Features & Real Ret. & Inpainting Int. & T2I Int. & Overall Acc. \\
        \hline
        The Bucket Effect & \(\mathbf{0.883}\) & 0.321 & 0.561 & 0.588 \\
        Analysis of Variance & 0.862 & 0.300 & 0.492 & 0.551 \\
        Skewness Analysis & 0.644 & 0.390 & 0.485 & 0.506 \\
        Gap Analysis & 0.768 & \(\mathbf{0.495}\) & \(\mathbf{0.762}\) & \(\mathbf{0.675}\) \\
        \hline
    \end{tabular}
\end{table}

Table~\ref{tab:single_feature} reports the univariate discrimination results. Among all features, the macro--micro gap \(F_{\mathrm{gap}}\) achieves the best overall performance, suggesting that AI images may preserve global lighting plausibility while showing weaker local detail coherence. \(F_{\min}\) and \(F_{\mathrm{var}}\) retain more Real samples but provide weaker AI interception. However, all single features remain limited on Inpainting, with the best interception rate only 49.5\%, indicating that local AI traces can be masked by the largely preserved real-image structure and motivating joint multidimensional classification.

\subsubsection{Multi-dimensional Feature Joint Discrimination.}
A single threshold cannot fully capture complex artifact patterns. Therefore, we concatenate the five original scores and four derived features into a \(9\)-dimensional vector and train a Random Forest classifier with \(5\)-fold stratified cross-validation. The results are shown in Table~\ref{tab:random_forest}.

\begin{table}
    \centering
    \caption{Results of \(5\)-fold cross-validation based on Random Forest.}
    \label{tab:random_forest}
    \resizebox{\textwidth}{!}{
    \begin{tabular}{lcccc}
        \hline
        Scoring Models & Real Ret. & Inpainting Int. & T2I Int. & Overall Acc. \\
        \hline
        GPT-\(5.2\) & \(0.578 \pm 0.018\) & \(0.465 \pm 0.011\) & \(0.706 \pm 0.015\) & \(0.588 \pm 0.011\) \\
        Gemini-\(2.5\)-Pro & \(\mathbf{0.841} \pm \mathbf{0.014}\) & \(0.428 \pm 0.021\) & \(0.838 \pm 0.014\) & \(\mathbf{0.702} \pm \mathbf{0.007}\) \\
        Doubao-\(1.5\)-Vision-Pro & \(0.252 \pm 0.009\) & \(0.754 \pm 0.014\) & \(\mathbf{0.927} \pm \mathbf{0.009}\) & \(0.644 \pm 0.003\) \\
        Qwen3-VL-235B & \(0.572 \pm 0.014\) & \(0.557 \pm 0.017\) & \(0.788 \pm 0.015\) & \(0.639 \pm 0.007\) \\
        Qwen3-VL-30B & \(0.612 \pm 0.021\) & \(0.534 \pm 0.030\) & \(0.783 \pm 0.032\) & \(0.643 \pm 0.015\) \\
        Llama-\(4\)-Maverick & \(0.459 \pm 0.021\) & \(0.599 \pm 0.026\) & \(0.764 \pm 0.012\) & \(0.607 \pm 0.009\) \\
        Llama-\(4\)-Scout & \(0.291 \pm 0.007\) & \(\mathbf{0.778} \pm \mathbf{0.009}\) & \(0.819 \pm 0.010\) & \(0.630 \pm 0.006\) \\
        \hline
    \end{tabular}
    }
\end{table}

Table~\ref{tab:random_forest} shows that joint features improve over single-feature thresholding, with Gemini-\(2.5\)-Pro achieving the best overall accuracy, increasing from 67.5\% to 70.2\%, mainly due to its high Real retention and strong Text2Image interception. Doubao-\(1.5\)-Vision-Pro and Llama-\(4\)-Scout obtain higher AI interception but much lower Real retention, indicating stronger AI-biased tendencies. Across models, Inpainting remains harder than Text2Image, suggesting that global statistical cues alone are insufficient for subtle local edits.

\subsection{Semantic Reasoning Experiment Based on Multimodal Large Models}

In AniPrO-MFR-Bench, we fix GLM-\(4.5\)V as the multimodal judge to evaluate how structured descriptions generated by different annotation models affect provenance reasoning. For each sample, GLM-\(4.5\)V receives the original image together with the five-dimensional structured descriptions produced by the model listed in the first column of Table~\ref{tab:mfr_bench}, and then predicts whether the image is \texttt{Real} or \texttt{AI}.

\begin{table}
    \centering
    \caption{Results of AniPrO-MFR-Bench with GLM-\(4.5\)V as the fixed judge.}
    \label{tab:mfr_bench}
    \resizebox{\textwidth}{!}{
    \begin{tabular}{lcccc}
        \hline
        Description Extraction Models & Real Ret. & Inpainting Int. & T2I Int. & Overall Acc. \\
        \hline
        GPT-\(5.2\) & 0.485 & 0.626 & 0.835 & 0.649 \\
        Gemini-\(2.5\)-Pro & \(\mathbf{0.584}\) & 0.657 & 0.967 & \(\mathbf{0.738}\) \\
        Doubao-\(1.5\)-Vision-Pro & 0.481 & 0.642 & 0.920 & 0.684 \\
        Qwen3-VL-235B & 0.489 & 0.613 & 0.902 & 0.678 \\
        Qwen3-VL-30B & 0.410 & 0.696 & 0.929 & 0.665 \\
        Llama-\(4\)-Maverick & 0.479 & 0.663 & 0.934 & 0.695 \\
        Llama-\(4\)-Scout & 0.354 & \(\mathbf{0.795}\) & \(\mathbf{0.975}\) & 0.710 \\
        \hline
    \end{tabular}
    }
\end{table}

As shown in Table~\ref{tab:mfr_bench}, Text2Image samples are generally easier to identify. In contrast, Inpainting samples are more challenging because they preserve much of the original image structure while introducing only local AI-generated regions. This supports evaluating Inpainting as a separate subcategory rather than simply merging all AI samples.

Another important observation is the relatively low Real retention across models. This suggests that multimodal models tend to over-classify stylized anime images as \texttt{AI}, possibly because intentional artistic exaggerations, such as simplified shading, distorted perspective, or flat color blocks, can be confused with AI-generated artifacts. Therefore, anime image provenance requires not only artifact detection, but also a better understanding of human artistic intent and stylized visual conventions.

Under the fixed GLM-\(4.5\)V judge, different annotation models lead to noticeably different provenance results, suggesting that structured descriptions provide discriminative cues beyond the image input alone. Gemini-\(2.5\)-Pro descriptions achieve the best overall accuracy with a relatively balanced Real/AI trade-off, whereas Llama-\(4\)-Scout achieves the strongest AI interception but the lowest Real retention. This further indicates that reliable provenance detection should balance AI interception with false-positive control on human-drawn images.

\subsection{Ablation Study}

\subsubsection{Input Modality Ablation.}
To analyze the roles of visual input and semantic descriptions, we evaluate three settings: Image-Only, Description-Only, and Ours. They use the image alone, Gemini-extracted descriptions alone, and both inputs, respectively. The results are shown in Table~\ref{tab:input_ablation}.

\begin{table}
    \centering
    \caption{Results of multimodal input ablation experiments. All descriptive texts are extracted uniformly by Gemini-\(2.5\)-Pro.}
    \label{tab:input_ablation}
    \resizebox{\textwidth}{!}{
    \begin{tabular}{llcccc}
        \hline
        Judge Models & Experimental Setup & Real Ret. & Inpainting Int. & T2I Int. & Overall Acc. \\
        \hline
        Gemini-\(2.5\)-Pro & Image-Only & \(\mathbf{0.986}\) & 0.181 & 0.968 & 0.807 \\
        Gemini-\(2.5\)-Pro & Ours & 0.938 & \(\mathbf{0.501}\) & \(\mathbf{0.970}\) & \(\mathbf{0.825}\) \\
        \hline
        GLM-\(4.5\)V & Image-Only & \(\mathbf{0.817}\) & 0.296 & 0.697 & 0.585 \\
        GLM-\(4.5\)V & Description-Only & 0.630 & 0.522 & 0.838 & 0.663 \\
        GLM-\(4.5\)V & Ours & 0.584 & \(\mathbf{0.657}\) & \(\mathbf{0.967}\) & \(\mathbf{0.738}\) \\
        \hline
    \end{tabular}
    }
\end{table}

Table~\ref{tab:input_ablation} shows that structured descriptions improve Inpainting detection in both same-model and cross-model settings. With Gemini-\(2.5\)-Pro extracting descriptions and judging provenance, adding descriptions increases Inpainting interception from 18.1\% to 50.1\%, maintaining high Real retention. For GLM-\(4.5\)V, where Gemini-extracted descriptions serve as external semantic evidence, Description-Only already outperforms Image-Only in overall accuracy, and the full multimodal setting further improves overall accuracy to 73.8\% and Inpainting interception to 65.7\%. These results indicate that five-dimensional descriptions provide transferable provenance cues complementary to visual perception.

\subsubsection{Dimension-level Ablation.}
To further examine the necessity of each semantic dimension, we conduct a leave-one-dimension-out ablation study. The results are shown in Table~\ref{tab:dimension_ablation}.

\begin{table}
    \centering
    \small
    \caption{Leave-one-dimension-out ablation results.}
    \label{tab:dimension_ablation}
    \resizebox{\textwidth}{!}{
    \begin{tabular}{lccccc}
        \hline
        Setting & Real Ret. & Inpainting Int. & T2I Int. & Overall Acc. & \(\Delta\)Acc. \\
        \hline
        Full 5D & 0.584 & 0.657 & \(\mathbf{0.967}\) & \(\mathbf{0.738}\) & -- \\
        w/o 3D Structures and Spatial Relationships & 0.588 & 0.594 & 0.871 & 0.693 & \(-0.045\) \\
        w/o Drawing Logic and Manual Control & 0.433 & \(\mathbf{0.704}\) & 0.894 & 0.678 & \(-0.060\) \\
        w/o Lighting and Textures & 0.651 & 0.604 & 0.944 & 0.712 & \(-0.026\) \\
        w/o Fine Details & 0.580 & 0.578 & 0.897 & 0.673 & \(-0.065\) \\
        w/o Contour and Line Quality & \(\mathbf{0.653}\) & 0.556 & 0.804 & 0.704 & \(-0.034\) \\
        \hline
    \end{tabular}
    }
\end{table}

As shown in Table~\ref{tab:dimension_ablation}, removing any dimension reduces overall accuracy, confirming the complementary roles of the five provenance cues. Removing Fine Details causes the largest accuracy drop, while removing Contour and Line Quality most severely reduces Inpainting interception. The decline in Real retention after removing Drawing Logic and Manual Control further suggests its role in distinguishing intentional human stylization from AI-generated artifacts. Single-dimension-only ablations are provided in the supplementary material.

\subsection{Additional Baseline Comparison}

To contextualize AniPrO, we evaluate two supervised visual classifiers and a small-scale human rater baseline. CNN and ViT-B/16~\cite{vit} are tested on the full AniPrO benchmark, while five anime enthusiasts classify a random 10\% subset.

\begin{table}
\centering
\caption{Supervised visual classifiers and human rater baselines. Human results are averaged over five anime enthusiasts on a randomly sampled 10\% subset.}
\label{tab:baseline_comparison}
\begin{tabular}{lcccc}
\hline
Method & Real Ret. & Inpainting Int. & T2I Int. & Overall Acc. \\
\hline
CNN baseline & 0.522 & 0.656 & 0.982 & 0.720 \\
ViT-B/16 & 0.835 & 0.211 & 0.551 & 0.532 \\
Human raters & 0.619 & 0.479 & 0.833 & 0.644 \\
\hline
\end{tabular}
\end{table}

Table~\ref{tab:baseline_comparison} shows that CNN favors AI interception, whereas ViT-B/16 is more conservative and misses many AI samples. Human raters retain more Real images than CNN but still struggle with Inpainting, indicating that anime provenance judgment remains difficult even for anime-familiar users.

\section{Conclusion}

We presented AniPrO, a balanced anime image provenance benchmark with five-dimensional semantic annotations, and a dual-track evaluation framework combining statistical feature discrimination with multimodal fusion reasoning. Statistically, the macro--micro score gap reveals a systematic discrepancy between global visual plausibility and local detail coherence in AI-generated anime images. Semantically, structured descriptions provide complementary provenance cues and improve the detection of challenging inpainting samples over image-only judging in our ablation setting. Meanwhile, the moderate overall accuracy and relatively low Real Retention show that anime provenance detection remains difficult, especially for highly stylized human-drawn artworks. Future work will extend AniPrO to more generators and artistic styles, and explore scalable deployment with lower-cost open-source models. Overall, interpretable visual decomposition provides a useful diagnostic foundation for AIGC copyright attribution, platform governance, and multimodal provenance research.

%
%
%
\bibliographystyle{splncs04}
\bibliography{paper276}

\end{document}